\documentclass[letterpaper, 10 pt, conference]{ieeeconf}  

\IEEEoverridecommandlockouts                              
\usepackage[table]{xcolor}
\usepackage{textcomp}
\usepackage{amssymb,amsmath,amsfonts}
\usepackage{graphicx}
\usepackage{tikz}
\usepackage{url}

\usepackage{colortbl}
\usepackage{booktabs}
\usepackage{array}
\usepackage{tabularx}

\newcolumntype{L}[1]{>{\raggedright\let\newline\\\arraybackslash\hspace{0pt}}m{#1}}

\newcommand{\mc}[1]{\ensuremath{\mathcal{#1}}}   

\newcommand{\nth}[1]{\ensuremath{#1^\text{th}}}

\makeatletter
\DeclareRobustCommand\onedot{\futurelet\@let@token\@onedot}
\def\@onedot{\ifx\@let@token.\else.\null\fi\xspace}

\usepackage[T1]{fontenc}  
\usepackage[b]{esvect}    

\makeatletter
\newlength\xvec@height%
\newlength\xvec@depth%
\newlength\xvec@width%
\newcommand{\xvec}[2][]{%
  \ifmmode%
    \settoheight{\xvec@height}{$#2$}%
    \settodepth{\xvec@depth}{$#2$}%
    \settowidth{\xvec@width}{$#2$}%
  \else%
    \settoheight{\xvec@height}{#2}%
    \settodepth{\xvec@depth}{#2}%
    \settowidth{\xvec@width}{#2}%
  \fi%
  \def\xvec@arg{#1}%
  \def\xvec@dd{:}%
  \def\xvec@d{.}%
  \raisebox{.2ex}{\raisebox{\xvec@height}{\rlap{%
    \kern.05em
    \begin{tikzpicture}[scale=1]
    \pgfsetroundcap
    \draw (.05em,0)--(\xvec@width-.05em,0);
    \draw (\xvec@width-.05em,0)--(\xvec@width-.15em, .075em);
    \draw (\xvec@width-.05em,0)--(\xvec@width-.15em,-.075em);
    \ifx\xvec@arg\xvec@d%
      \fill(\xvec@width*.45,.5ex) circle (.5pt);%
    \else\ifx\xvec@arg\xvec@dd%
      \fill(\xvec@width*.30,.5ex) circle (.5pt);%
      \fill(\xvec@width*.65,.5ex) circle (.5pt);%
    \fi\fi%
    \end{tikzpicture}%
  }}}%
  #2%
}
\makeatother

\makeatletter
\renewcommand*\env@matrix[1][\arraystretch]{%
  \edef\arraystretch{#1}%
  \hskip -\arraycolsep
  \let\@ifnextchar\new@ifnextchar
  \array{*\c@MaxMatrixCols c}}
\makeatother

\newcommand{\FrameSymbol}{\mc{F}}                       
\newcommand{\frm}[1]{\mc{#1}}                           
\newcommand{\Frm}[1]{{{\FrameSymbol}^{\frm{#1}}}}       
\newcommand{\mat}[1]{{\mathbf{#1}}}                     
\newcommand{\point}[1]{{#1}}

\newcommand{\unit}[1]{$\left[ #1 \right]$}

\definecolor{commentcolor}{gray}{0.5}
\usepackage{algorithm}
\usepackage{algpseudocode}

\algnewcommand{\LineComment}[1]{\State \textcolor{commentcolor}{\(\triangleright\) #1}}
\algnewcommand{\To}{\textbf{to}}
\algnewcommand{\Break}{\textbf{break}}
\algnewcommand{\Continue}{\textbf{continue}}
\algnewcommand{\IIf}[1]{\State\algorithmicif\ #1\ \algorithmicthen}
\algnewcommand{\EndIIf}{\unskip}
\algnewcommand{\var}[1]{\textit{#1}}
\algnewcommand{\func}[1]{\textsc{#1}}

\newcommand{\FI}{\Frm{I}}
\newcommand{\FB}{\Frm{B}}

\newcommand{\FW}{\Frm{W}}

\newcommand{\OI}{\point{O}_{\frm{I}}}
\newcommand{\OB}{\point{O}_{\frm{B}}}
\newcommand{\OW}{\point{O}_{\frm{W}}}

\newcommand{\XI}{\hat{e}_n}
\newcommand{\YI}{\hat{e}_e}
\newcommand{\ZI}{\hat{e}_d}
\newcommand{\XB}{\hat{e}_x}
\newcommand{\YB}{\hat{e}_y}
\newcommand{\ZB}{\hat{e}_z}
\newcommand{\XW}{\hat{x}_w}
\newcommand{\YW}{\hat{y}_w}
\newcommand{\ZW}{\hat{z}_w}
\newcommand{\ZR}[1]{\hat{u}_{#1}}

\newcommand{\rB}[1]{\mathbf{r}_#1^\frm{B}}

\newcommand{\Force}[1]{\mat{F}^\frm{B}_{#1}}
\newcommand{\Moment}[1]{\mat{M}^\frm{B}_{#1}}
\newcommand{\Thrust}[1]{\mat{T}_{#1}}
\newcommand{\Torque}[1]{\mat{\tau}_{#1}}

\newcommand{\RBI}{\mat{R}^{\frm{B}}_{\frm{I}}}
\newcommand{\RIB}{\mat{R}^{\frm{I}}_{\frm{B}}}
\newcommand{\RBW}{\mat{R}^{\frm{B}}_{\frm{W}}}

\newcommand{\airlab}{\url{https://theairlab.org/vtol}}

\def\publishedurl{\centering \textbf{{\color{red}{The published version can be accessed from}} \color{blue}{\url{https://ieeexplore.ieee.org/document/9981806}}}}

\usepackage{cite}
\usepackage{soul}
\usepackage{environ}
\usepackage{multirow}
\usepackage[table]{xcolor}
\usepackage{url}
\usepackage{color}

\usepackage{mathtools}
\usepackage{siunitx}
\usepackage{adjustbox}
\usepackage{float}
\usepackage[font=small,tableposition=top]{caption}
\usepackage{booktabs}

\usepackage{pgfplots}
\pgfplotsset{compat=newest} 
\usetikzlibrary{plotmarks}
\usepgfplotslibrary{patchplots}
\usepackage{grffile}
\usepackage{threeparttable}
\usepackage{gensymb}

\title{\LARGE \bf
Design, Modeling and Control for a Tilt-rotor VTOL UAV \\in the Presence of Actuator Failure
}

\author{Mohammadreza Mousaei$^{1}$, Junyi Geng$^{1}$, Azarakhsh Keipour$^{2}$, Dongwei Bai$^{1}$, and Sebastian Scherer$^{1}$
\thanks{$^{1}$ The Robotics Institute, Carnegie Mellon University, Pittsburgh, PA 15213, USA.
        {\tt\small \{mmousaei, junyigen, dongweib, basti\}@andrew.cmu.edu}}%
\thanks{$^{2}$ Robotics AI, Amazon, Washington, DC 20009, USA.
        {\tt\small keipour@gmail.com}}%
\thanks{* During the realization of this work, A. Keipour was affiliated with Carnegie Mellon University. The publication was written prior to A. Keipour joining Amazon.}%
    }

\begin{document}

\maketitle
\thispagestyle{empty}
\pagestyle{empty}

\begin{abstract}

Enabling vertical take-off and landing while providing the ability to fly long ranges opens the door to a wide range of new real-world aircraft applications while improving many existing tasks. Tiltrotor vertical take-off and landing (VTOL) unmanned aerial vehicles (UAVs) are a better choice than fixed-wing and multirotor aircraft for such applications. Prior works on these aircraft have addressed the aerodynamic performance, design, modeling, and control. However, a less explored area is the study of their potential fault tolerance due to their inherent redundancy, which allows them to tolerate some degree of actuation failure. This paper introduces tolerance to several types of actuator failures in a tiltrotor VTOL aircraft. We discuss the design and modeling of a custom tiltrotor VTOL UAV, which is a combination of a fixed-wing aircraft and a quadrotor with tilting rotors, where the four propellers can be rotated individually. Then, we analyze the feasible wrench space the vehicle can generate and design the dynamic control allocation so that the system can adapt to actuator failures, benefiting from the configuration redundancy. The proposed approach is lightweight and is implemented as an extension to an already-existing flight control stack. Extensive experiments validate that the system can maintain the controlled flight under different actuator failures. To the best of our knowledge, this work is the first study of the tiltrotor VTOL's fault-tolerance that exploits the configuration redundancy. The source code and simulation can be accessed from \airlab.

\end{abstract}


\section{Introduction} \label{sec:intro}

Unmanned Aerial Vehicles (UAVs) have gained interest in various applications, ranging from 3D mapping and photography~\cite{zhao2021super} to aerial manipulation and physical interaction\cite{geng2020cooperative, Keipour:2020:arxiv:integration}. UAVs with fixed rotors (e.g., multirotors) have hover and vertical take-off and landing (VTOL) capabilities~\cite{Keipour:2022:sensors:mbzirc}. However, due to the significant upward thrust required to counter the gravity, they are inefficient in the forward flight. On the other hand, fixed-wing vehicles (e.g., airplanes) are very efficient in forward flight and can fly much longer ranges than multirotors. However, they lack VTOL and hover capabilities and usually require runways for take-off and landings. As the third class, VTOL hybrid or convertible UAVs combine VTOL capabilities with efficient forward flight by using propellers to hover, take-off, and land vertically, and wings for efficient long-range cruising. 

In general, there are three main types of VTOLs: tailsitters, tiltrotors, and standard VTOLs. However, it is challenging to balance the configuration complexity and the control simplicity. For example, although standard VTOLs are the easiest to control, they add additional weight from separate hover/forward flight propulsion systems. Tailsitter VTOLs have the minimum set of actuators; however, they can be hard to control, particularly in the wind. As for the tiltrotors, they are easier to control in hover than tailsitters due to more control authority, but the additional tilting mechanism increases the system's complexity. 

\begin{figure}
    \centering
    \includegraphics[width=0.48\textwidth]{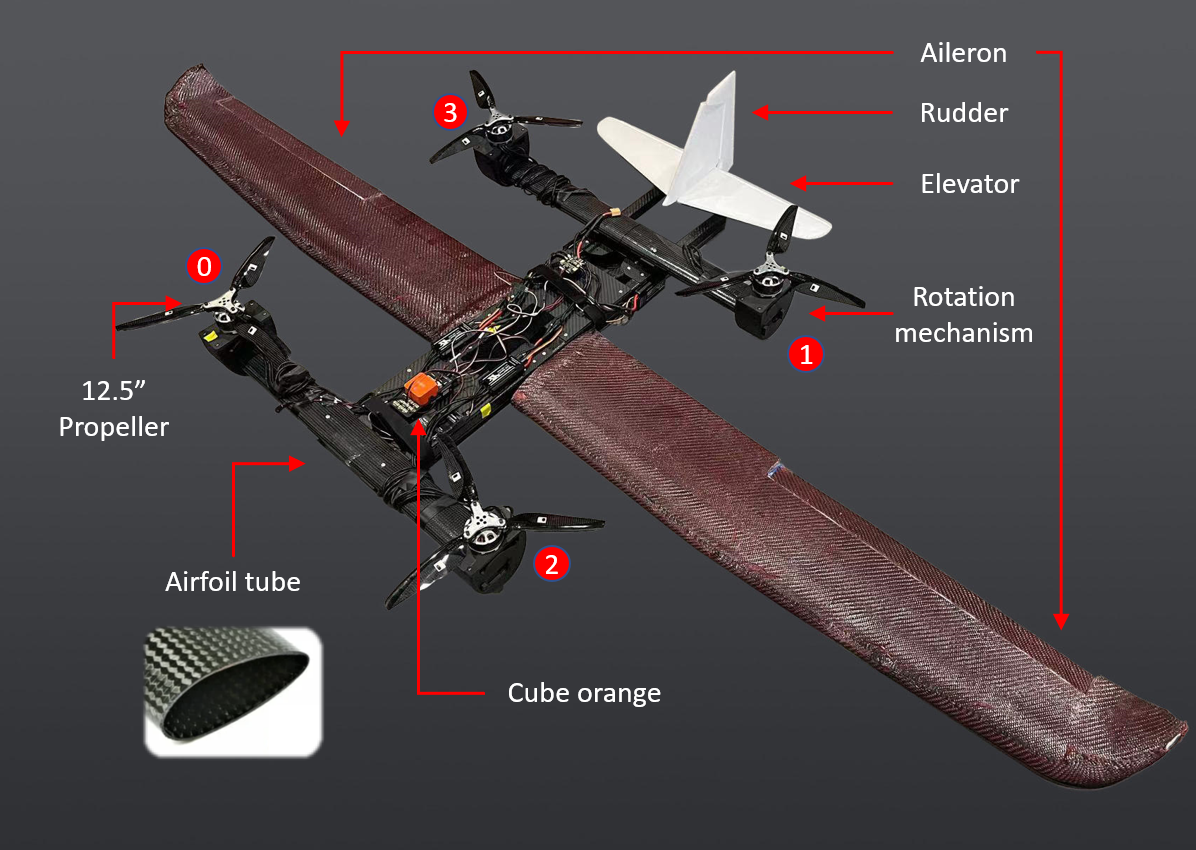}
    \caption{Tiltrotor VTOL aircraft designed in this work with four rotors, four tilting mechanisms, two ailerons, one elevator, and one rudder. The quadrotor arms have airfoil cross-sections.}
    \label{fig:vtol-title}
    \vspace{-3mm}
\end{figure}

Safety requirements have become more and more critical as the number of operating UAVs grow. To safely integrate UAVs into the airspace and for real-world applications, the aircraft should maintain controlled flight even when actuator failure happens, with a subsequent fail-safe mechanism, i.e., return-to-home or emergency landing. Hence, the flight control system must be able to tolerate some degree of failure. In general, such kind of control system with the fault-tolerant ability can either be achieved by designing the controller in a robust manner~\cite{robust1, robust2} or by dynamically adapting the controller to the detected failure. While the former usually requires a complete redesign based on the aircraft configuration, the latter requires additional fault detection and identification. 
This paper focuses on the latter option, assuming that the system failure can be quickly detected and identified so that the existing autopilots with more powerful flight stacks can be used with a minimum amount of further development. Real-time failure detection methods for aircraft already exist that can provide the failure status almost immediately after the failure happens~\cite{Keipour:2019:icra:anomaly, Mousaei:2022:icra-workshop:vtol}.

There are many ways to handle actuator failures depending on the aircraft's configuration. Multirotors with less than six propellers are still able to fly after the loss of one or more propellers~\cite{mueller2016relaxed}. However, with a failed motor they cannot statically hover and can only keep the position while rotating in-place around the gravity vector. This concept of dynamic hovering can be an acceptable compromise if the flying space is not constrained and if the perception system can still operate in such conditions. It has been shown that six rotors are required at minimum to achieve static hovering robustness to a motor failure~\cite{michieletto2018fundamental}. In particular, only star-shaped hexarotors with tilted propellers and Y-shaped hexarotors can achieve static hovering robustness~\cite{baskaya2021novel}. On fixed-wing aircraft, exploiting aerodynamic effects for passive stability or utilizing actuator redundancy for failure adaptation are the possible ways to achieve fault-tolerance~\cite{stastny2018nonlinear}. As for the hybrid VTOL UAVs, although the existing works addressed the aerodynamic performance~\cite{busan2021wind}, design~\cite{lyu2017design, kamal2020conceptual}, modeling~\cite{ducard2014modeling} and control~\cite{zhang2018control, bauersfeld2021mpc}, fault-tolerance is less explored. Due to the hybrid configuration design, VTOLs have configuration redundancy. However, the research utilizing redundancy to address actuator failures is minimal. The authors of~\cite{fuhrer2019fault} investigate the fault-tolerant flight control of a tailsitter VTOL. Aerodynamic passive stability and control allocation adjustment are exploited to achieve the fault-tolerant system performance. 

This paper introduces tolerance to several types of actuator failures in a tiltrotor VTOL aircraft. The model and design of a custom tiltrotor VTOL are discussed, which combines a fixed-wing aircraft and a variable-pitch quadrotor UAV. The direction of the four propellers can be individually controlled. Different from the tilting design of~\cite{ducard2014modeling} with an additional arm length, twin-rotors~\cite{papachristos2011design} or other tilting multi-rotors~\cite{hua2014control}, our quadrotor arms are entirely separated from the main wing, which makes the control less affected by the unexpected wing deformation. In addition, our tilting mechanism directly tilts the rotors at the end of the quadrotor arm. Moreover, the quadrotor arms have airfoil cross-sections, which can provide additional lift for the vehicle during forward motion. We model the nonlinear dynamics of this tiltrotor VTOL and analyze the feasible wrench space the vehicle can generate. Then, we design the dynamic control allocation that allows the system to adapt to the potential configuration changes in real-time. This dynamic control allocation benefits from configuration redundancy to make the aircraft robust to actuator failures. By solving a constraint optimization problem under a carefully designed objective, the aircraft can recover from a set of actuator failures in different flight phases of the VTOL.

Our main contributions include:
\begin{enumerate}
    \item Proposing a dynamic control allocation method that allows the system to adapt to actuator failures. The proposed approach is light-weight and can be quickly extended on an already-existing flight control stack;
    \item Designing and modeling a tiltrotor VTOL with the ability to rotate each individual propeller; 
    \item Validating the system performance under the set of possible actuator failures in different flight phases;
    \item Providing the source code for the proposed strategies implemented on the PX4 flight controller firmware along with our simulation environment.
\end{enumerate}

\section{System Overview} \label{sec:system}

This section presents the designed aircraft and its avionics system.

\subsection{Vehicle Design}

\begin{figure}
    \centering
    \includegraphics[width=\linewidth]{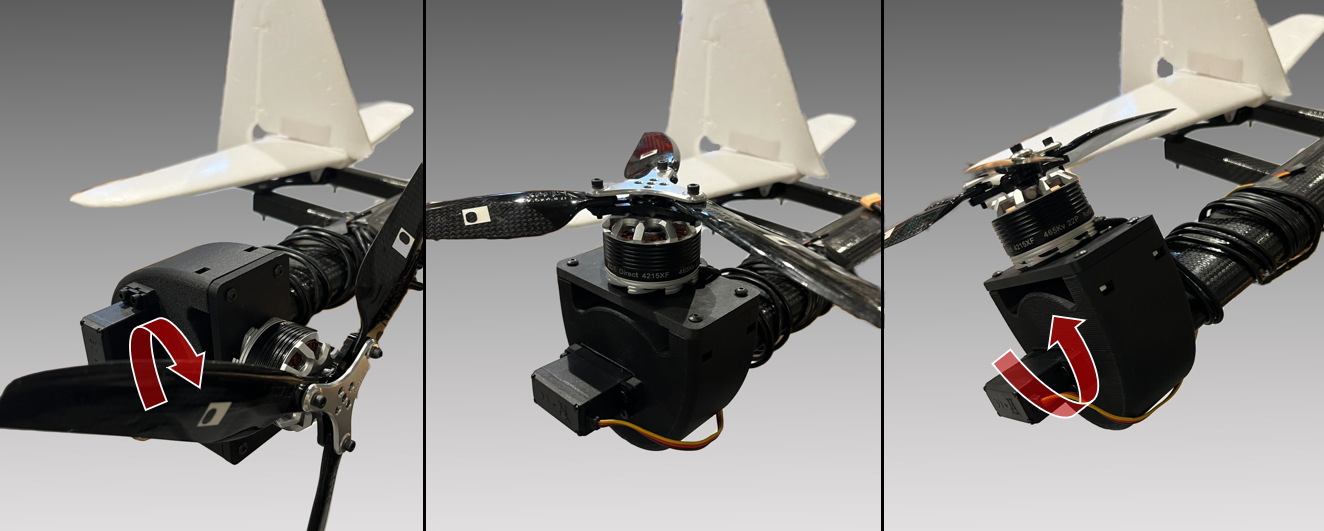}
    \caption{The rotation mechanism in three different positions.}
    \label{fig:tilt}
    \vspace{-3mm}
\end{figure}

The aircraft has two streamlined airfoil tubes integrated vertically into the fuselage as the quadrotor arms. Different from the design of~\cite{ducard2014modeling} that has the quadrotor arms parallel to the fuselage, our design makes the arm completely separate from the main wing. This provides several advantages: (1) the control authority of the tiltrotor gets less affected by the unexpected wing structure deformation; (2) the tilt mechanism can be directly attached to the end of the arm without a conflict with the structure; (3) the airfoil cross-section of the arm provides additional lift to the aircraft. We attach four brushless electrical motors directly to the arms' ends, where they can drive a 12.5" propeller and provide 27.36~\unit{N} at 100\% throttle. The component that connects the motor and propeller to the fuselage is the rotation mechanism powered by an MG996R servo which has a maximum rotation angle of 180\textdegree\ and enables the propeller to tilt forward and backward, as shown in Figure~\ref{fig:tilt}. 

A set of wings with a wingspan of 2~\unit{m} is attached to the fuselage. The elevator and rudder are mounted to the extended structure from the fuselage. A 16000~mAh 6S LiPo battery is mounted on the fuselage. Figure~\ref{fig:vtol-title} shows the aircraft and its avionics.

\subsection{Avionics}
A cube orange flight controller~\cite{HpJetFusion} is the core of the avionics. It features three inertial measurement units (IMUs), an integrated accelerometer, a barometer, a gyroscope, and a 400~Mhz 32-bit ARM Cortex-M7 processor. A long-range telemetry, a differential pitot tube, and a GPS are connected to the cube orange to provide communication, airspeed, and position updates. The Extended Kalman Filter (EKF) of the PX4 open source firmware~\cite{meier2015px4} performs sensor fusion to estimate the aircraft states and enable automatic control. 


\section{Model} \label{sec:model}

This section briefly describes the definitions and the mathematical model of our VTOL system used for control. A more detailed model is discussed in~\cite{ducard2014modeling}.

\subsection{Definitions}

The inertial frame is defined as $\FI=\{\OI, \XI, \YI, \ZI\}$, where $\OI$ is the 3-D origin point, and $\XI$, $\YI$, and $\ZI$ are the unit vectors pointing to the North, East and Down directions. The body-fixed frame is defined as $\FB=\{\OB, \XB, \YB, \ZB\}$, where $\OB={\mathbf{p}}$ is the position of the vehicle's center of mass in the inertial frame, and $\XB$, $\YB$, and $\ZB$ are the unit vectors pointing to the front, right and bottom directions of the vehicle, respectively. The wind frame is defined as $\FW=\{\OW, \XW, \YW, \ZW\}$ where $\OW$ is arbitrary, the $\XW$ axis is in the positive direction of the vehicle's velocity relative to the wind, $\ZW$ is perpendicular to $\XW$ and positive below the vehicle, and $\YW$ is perpendicular to the $\XW\ZW$ plane following the right-hand rule. The angle of attack $\alpha$ is the angle of the velocity vector with the $\XW\YW$ plane, and the sideslip angle $\beta$ is the angle between the velocity vector and the projection of the vehicle's longitudinal axis to the $\XW\YW$ plane. $\RBI$ and $\RIB$ define the rotations from $\FB$ to $\FI$ and $\FI$ to $\FB$, respectively.

As described in Section~\ref{sec:system}, the system has 12 actuators that can be described using a set of parameters. The deviation angles of the left and right ailerons from the neutral position are $\delta_{a_1}$ and $\delta_{a_2}$, respectively. The elevator and rudder deviation angles from their neutral positions are described by $\delta_e$ and $\delta_r$. The rotation velocity of the \nth{i} rotor is represented by $\omega_i$. Finally, the rotation angle for each rotor $i$ is described by $\chi_i$. Angle $\chi_i$ is designated as the rotation angle of the \nth{i} rotor's axis with $-\ZB$ around the $-\YB$ axis. Therefore, $\chi_i$ will be zero when the \nth{i} rotor is pointing upward and $\frac{\pi}{2}$ when they are pointing to the front of the vehicle. All angles are described in \textit{radians}.

Finally, the thrust and torque generated by the $\nth{i}$ rotor are defined as $\Thrust{i}$ and $\Torque{i}$, the unit vector in the direction of $\Thrust{i}$ (rotor axis) is $\ZR{i}$ and the vector from the body-fixed origin (i.e., the center of mass) to the base of the $\nth{i}$ rotor arm is defined as $\rB{i}$. Assuming that all rotors are similar, the magnitude of the thrust $\Thrust{i}$ and torque $\Torque{i}$ can be approximated as:
\begin{align}
\begin{split}
    &\|\Thrust{i}\| = T_i = c_F \omega_i^2\\
    &\|\Torque{i}\| = \tau_i = (-1)^{d_i} c_K \omega_i^2
\end{split}
\end{align}
\noindent where $c_F > 0$ and $c_K > 0$ are the thrust and torque constant coefficients of the rotors, and $d_i = \{0, 1\}$ is the rotation direction of the \nth{i} rotor around its axis (i.e., clockwise or counter-clockwise).

\subsection{Forces}

Three main forces are applied to the VTOL: gravitational, thrust, and aerodynamic. We describe these forces in the body-fixed frame $\FB$ for convenience.

Given the total mass of the vehicle as $m$, the gravitational force is:
\begin{equation}
    \Force{g} = \RBI\begin{bmatrix}0\\0\\mg\ZI\end{bmatrix}
\end{equation}

The thrust forces are the result of the thrusts generated by the propellers. Assuming a negligible induced drag, for the total thrust force we have:
\begin{equation}
    \Force{r} = \sum_{i = 1}^4 \Thrust{i}^\frm{B} =
    \sum_{i = 1}^4 T_i\ZR{i}^\frm{B} =
    c_F\sum_{i = 1}^4 \omega_i^2 \begin{bmatrix}\sin {\chi_i}\\0\\-\cos{\chi_i}\end{bmatrix}
\end{equation}

The aerodynamic forces are the forces exerted on the robot by the air and can be expressed as:
\begin{equation}
    \Force{a} = \RBW\begin{bmatrix}X^\frm{W} \\Y^\frm{W} \\ Z^\frm{W}\end{bmatrix}
\end{equation}
\noindent where we have:
\begin{align}\label{eq:model:force-coeffs}
\begin{split}
    X^{\frm{W}} &= \Bar{q} S C_X(\alpha, \beta) ~~~\mbox{(Drag Force)}\\
    Y^{\frm{W}} &= \Bar{q} S C_Y(\beta) ~~~~~~\mbox{(Lateral Force)}\\
    Z^{\frm{W}} &= \Bar{q} S C_Z(\alpha) ~~~~~~\mbox{(Lift Force)} 
\end{split}
\end{align}

In this equation, $S$ is the wing surface area and $\Bar{q} = \rho V_a^2 / 2$ is the dynamic pressure, where $\rho$ is the air density and $V_a$ is the airspeed. The lateral force is negligible (i.e., $C_Y \approx 0$) during the typical cruise flight. For the lift and drag coefficients we have:
\begin{align}
\begin{split}
    &C_X(\alpha, \beta) \approx C_{D, 0} + C_{D, \alpha} \alpha^2 \\
    &C_Z(\alpha) \approx C_{Z, 0} + C_{Z, \alpha} \alpha
\end{split}
\end{align}
\noindent where $C_{D, 0}$ and $C_{D, \alpha}$ are the coefficients of parasite drag and induced drag, and $C_{Z, 0}$ and $C_{Z, \alpha}$ are the lift coefficients, and all are usually obtained from wind tunnel tests. 

The total forces applied to the VTOL are calculated as the sum of all three force types:
\begin{equation}\label{eq:totalF}
    \Force{} = \Force{r} + \Force{a} + \Force{g}
\end{equation}

\begin{figure*}[t]
    \centering
    \includegraphics[width=0.98\textwidth]{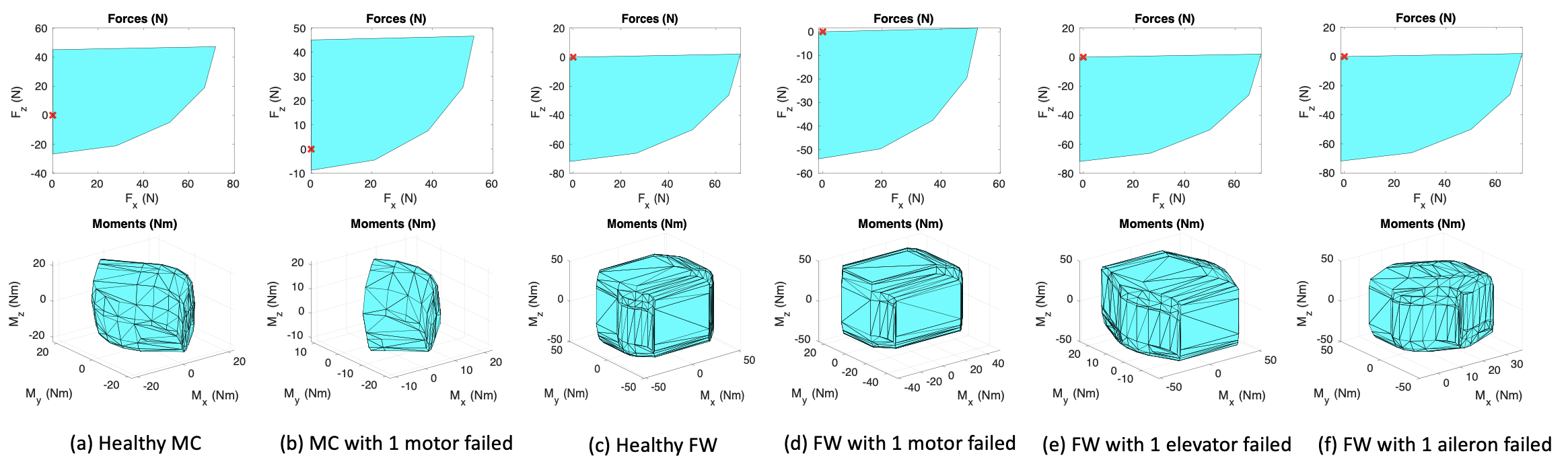}
    \caption{Visualization of the tiltrotor VTOL's feasible wrench sets. MC stands for multirotor and FW stands for the fixed-wing configuration. The wrench sets are computed around the trimmed condition for fixed-wing. Wrench analysis performed in the AirLab simulator using the method proposed in~\cite{Keipour:2022:thesis, Keipour:2023:scitech:simulator}.}
    \label{fig:ws_analysis}
    \vspace{-3mm}
\end{figure*}

\subsection{Moments}

There are several types of moments affecting the VTOL vehicle. However, in reality, the gyroscopic torques can be ignored for normal steady operations. Therefore, the principal moments applied to the VTOL are thrust, resisting, and fixed-wing aerodynamic moments.

Thrust moments are the result of the rotor thrust being applied at an offset from the center of mass. These moments can be computed as:
\begin{equation}
    \Moment{r} = \sum_{i = 1}^4 \left( \rB{i} \times \Thrust{i} \right)
\end{equation}

Resisting moments are the result of rotor rotation and rotor arm rotation applying the opposite rotation moments to the robot's body. If we assume the effect of the rotor arm rotation is small, the resisting moments can be computed as:
\begin{equation}
    \Moment{g} = \sum_{i = 1}^4 \Torque{i}^\frm{B} =
    c_K\sum_{i = 1}^4 (-1)^{d_i} \omega_i^2 \begin{bmatrix}\sin {\chi_i}\\0\\-\cos{\chi_i}\end{bmatrix}
\end{equation}

Aerodynamic moments are the moments exerted on the aircraft in fixed-wing configuration:
\begin{equation}
    \Moment{a} = \begin{bmatrix}L^\frm{B} \\ M^\frm{B} \\ N^\frm{B} \end{bmatrix}
\end{equation}

\noindent where for a cruise flight with small slide slip angle and angular rate, it can be approximated as:
\begin{align}
\begin{split}
    L^\frm{B} &\approx \Bar{q}SbC_{La}\delta_a \\
    M^\frm{B} &\approx \Bar{q}S\Bar{c}C_{Me}\delta_e \\
    N^\frm{B} &\approx \Bar{q}SbC_{Nr}\delta_r
\end{split}
\end{align}

In this equation, $S$ is the wing surface area, $\Bar{q}$ is the dynamic pressure, $\Bar{c}$ is the mean aerodynamic chord, and $b$ is the wingspan. $C_{La}$ is the effectiveness of the aileron, $C_{Me}$ is the effectiveness of the elevator, and $C_{Nr}$ is the effectiveness of the rudder. These coefficients are usually obtained from wind tunnel test data.

Finally, the total moment is calculated as:
\begin{equation}\label{eq:totalM}
    \Moment{} = \Moment{r} + \Moment{g} + \Moment{a}
\end{equation}


\section{Feasible Wrench Space}\label{sec:wrench-space}

This section aims to perform wrench space analysis for the tiltrotor VTOL and find their correspondence for different flight configurations.

From Equations~\ref{eq:totalF} and~\ref{eq:totalM}, it can be seen that generally the total wrench $\mathbf{W}^\mathcal{B} = [\mathbf{M}^\mathcal{B}, \mathbf{F}^\mathcal{B}]^{\top}$ is nonlinear with respect to the actuator inputs $\mathbf{u}$. In other words, $\mathbf{F}^\mathcal{B} = f(\mathbf{u})$ and $\mathbf{M}^\mathcal{B} = g(\mathbf{u})$ except in multirotor configuration, where $\mathbf{W}^\mathcal{B}$ becomes linear with respect to $\mathbf{u}$ (notated as $\mathbf{F}^\mathcal{B} = \mathbf{F}_m\mathbf{u}$ and $\mathbf{M}^\mathcal{B} = \mathbf{G}_m\mathbf{u}$). Let us assume that each entry of the input $\mathbf{u}$ is limited with the lower and upper bound to be $u_{\mathrm{min}}$ and $u_{\mathrm{max}}$, i.e., $\mathbf{u} \in \mathbb{U} = [u_{\min}, u_{\max}]$, where $\mathbb{U}$ is the set of feasible inputs. Then, the feasible wrench set is defined as the image set of $\mathbb{U}$ through the nonlinear map $h(\cdot)$:
\begin{equation}
    \mathcal{W} = \lbrace w \in \mathbb{R}^n~|~ \forall\mathbf{u} : w = h(\mathbf{u})\rbrace.
\end{equation}
Following this definition, the sets of feasible inputs at steady-state operation, as the sets of control inputs that can maintain the hover and cruise are defined as: 
\begin{align}
    \mathbb{U}_{h} &= \lbrace u \in \mathbb{R}^n~|~ \mathbf{u} : ||\mathbf{F}_m\mathbf{u}|| \geq mg\rbrace \\
    \mathbb{U}_{c} &= \lbrace u \in \mathbb{R}^n~|~ \mathbf{u} : f_x(\mathbf{u}) \geq 0 \rbrace
\end{align}
where $f_x(\cdot)$ is the x-component of the result vector. Feasible wrench sets at hover or cruise $\mathcal{W}_{\cdot +}$ are defined as the image set of $\mathbb{U}_{(\cdot)}$ through the mapping $h(\cdot)$. Note that $\mathcal{W}_{\cdot +} \in \mathcal{W}$ since $\mathbb{U}_{(\cdot)} \in \mathbb{U}$.

\subsection{Static Hovering}
The vehicle is capable of static hovering when it can reach and maintain a constant position and orientation, i.e., 
\begin{equation}
    \dot{\mathbf{p}}^{\mathcal{B}}\rightarrow\mathbf{0},\quad \boldsymbol{\omega}^{\mathcal{B}}\rightarrow\mathbf{0}
\end{equation}
Michieletto et al.~\cite{michieletto2018fundamental} have proven that the following conditions are required for keeping the static hovering ability:
\begin{align}
    \text{rank}\lbrace \mathbf{F}_m \rbrace &= 3 \\
    \exists \mathbf{u} \in \mathrm{int}(\mathbb{U}) & s.t. \left\{ \begin{array}{c}
         ||\mathbf{F}_m\mathbf{u}|| \geq mg  \\
         \mathbf{G}_m \mathbf{u} = \mathbf{0}
    \end{array} \right.
\end{align}
where $\mathrm{int}$ denotes the interior of $\mathbb{U}$.

It is clear that the aircraft can statically hover if and only if $\mathbf{0} \in \mathrm{int}(\mathcal{W}_{h+})$. If the origin is on the boundary or is outside of $\mathcal{W}_{h+}$, the aircraft cannot statically hover since the system will not be robust to disturbance. 

\subsection{Cruise Flight}

The vehicle is able to keep cruise flight in the fixed-wing configuration when it can keep the constant altitude, heading and forward speed, i.e.,
\begin{equation}
    \ddot{\mathbf{p}}_x^{\mathcal{B}}\rightarrow\mathbf{0},\quad \dot{\mathbf{p}}_z^{\mathcal{B}}\rightarrow\mathbf{0},\quad \dot{\psi} \rightarrow 0
\end{equation}
The wrench needs to satisfy the following condition to keep cruise flight:
\begin{equation}
    \exists \mathbf{u} \in \mathrm{int}(\mathbb{U}) \quad \& \quad s.t. \left\{ \begin{array}{c}
         f_x(\mathbf{u}) \geq 0  \\
         f_z(\mathbf{u}) \geq mg \\
         g(\mathbf{u}) = \mathbf{0}
    \end{array} \right.
\end{equation}

The VTOL's steady state ability can be seen from the feasible wrench set (see Figures~\ref{fig:ws_analysis}~(a) and~(c)). 

\subsection{Actuator Failures}

This section highlights the effect of an actuator failure on the VTOL's steady-state capability. Here, the actuator failure can be a motor failure, a servo tilt failure, or a control surface failure. We denote $\,^k \mathcal{W}_{\cdot +}$ the feasible wrench set at steady-state when the $k^{\mathrm{th}}$ actuator fails. 

Figures~\ref{fig:ws_analysis}(b) and~(d-f) show the wrench space when a motor fails at hover or cruise flight or a control surface fails at cruise flight.\footnote{~The wrench space for the tilt failure is not shown here due to its similarity to the motor failure.} We can see that a single motor failure causes a significant cutoff of the feasible wrench space for both the multirotor and fixed-wing configurations. The payload capacity was clearly reduced (see the force). When the elevator or the aileron fails, the range of $M_y$ or $M_x$ gets narrowed. In fact, the elevator is the main pitch ($\mathbf{M}_y$) control authority while the aileron controls the roll motion ($\mathbf{M}_x$).

\begin{figure}
    \centering
    \includegraphics[width=\linewidth]{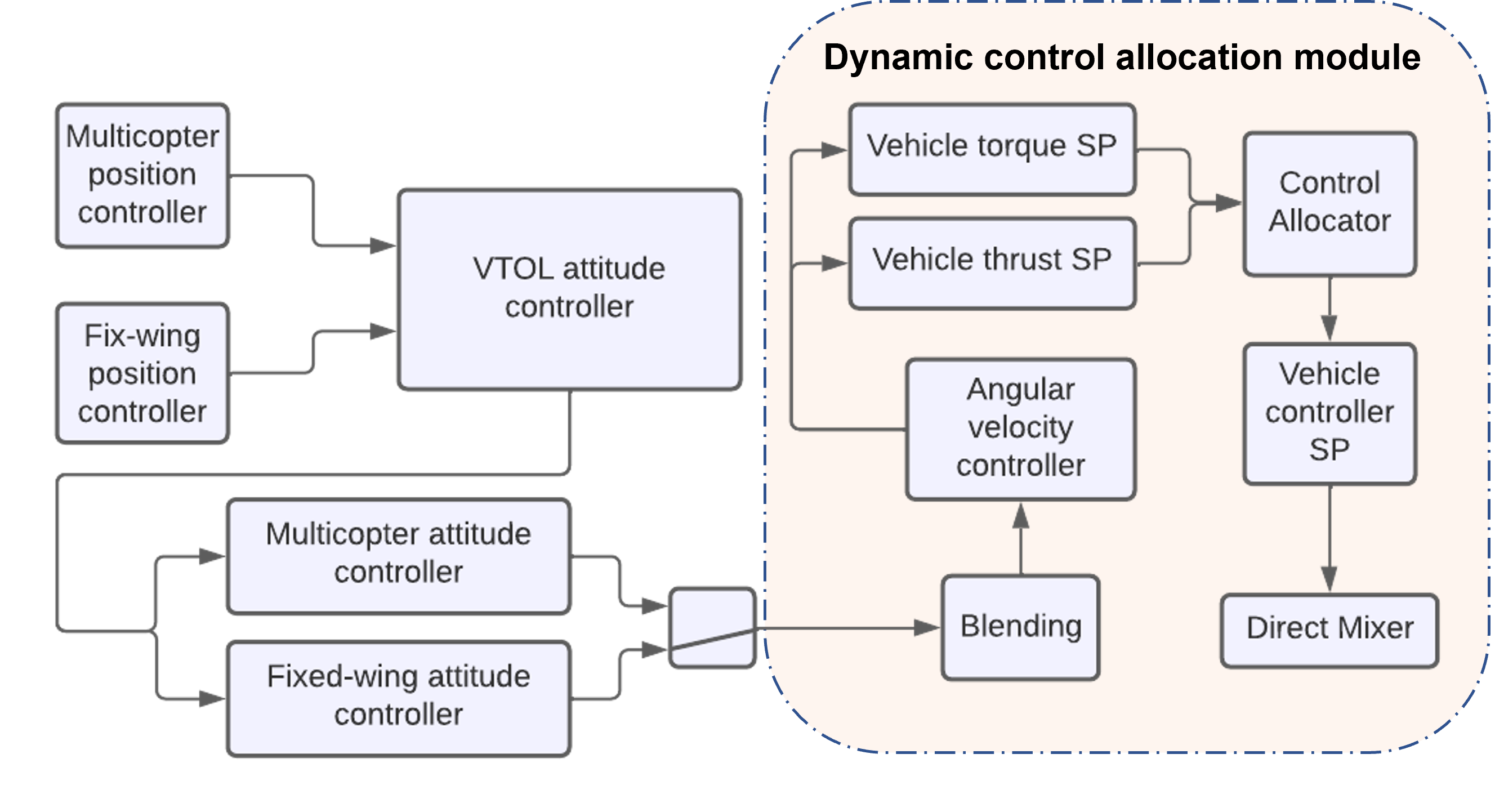}
    \caption{VTOL control diagram with dynamic control allocation.}
    \label{fig:diagram}
    \vspace{-3mm}
\end{figure}

\section{Nominal Control} \label{sec:nominal-control}
This section describes the actuation principles and the control diagram of the nominal flight case without any actuator failures. 

\subsection{Nominal Actuation Principle}
Our tiltrotor VTOL control scheme consists of a multirotor mode and a fixed-wing mode, either running separately in the corresponding VTOL phases or together during the transition. The vehicle starts with the multirotor phase, in which the UAV takes off vertically and keeps hovering at the desired altitude. Then, when a transition is triggered, the vehicle speeds up laterally and tilts the rotors to make them face forward. The transition phase ends when the vehicle gains enough airspeed and fully tilts the rotors. The vehicle subsequently stays in the fixed-wing phase for the rest of the flight until it transitions back to hovering and the multirotor phase for a vertical landing. Note that the specific VTOL attitude control facilitates the necessary switching and blending logic for multirotor and fixed-wing modes during the transition phase.  

\subsection{Dynamic Control Allocation}
Although PX4 provides the off-the-shelf controller for this type of aircraft, it usually needs to load the proper mixer files to map the desired control wrench to each actuator input, which can only handle the aircraft with a fixed configuration. Instead, we developed a dynamic control allocation scheme, so it adapts to a potential vehicle configuration change, such as an actuator failure. Therefore, the modified control diagram becomes like Figure~\ref{fig:diagram}. 

Given the desired control wrench $\mathbf{W}^\mathcal{B}$, i.e., the force and torque from the velocity and angular rate controllers, the goal is to allocate them to each of the actuators based on the current states and the aircraft configuration. We formulate a constrained optimization problem to perform this allocation. Specifically, in the transition and fixed-wing phases, we linearize $h(\cdot)$ with respect to the trimmed vehicle state based on the small perturbation theory. 
\begin{equation}
    \mathbf{F}^\mathcal{B} = \mathbf{F}_{\mathrm{trim}}^\mathcal{B} + \mathbf{F}_{\Delta}^\mathcal{B}, \quad
    \mathbf{M}^\mathcal{B} = \mathbf{M}_{\mathrm{trim}}^\mathcal{B} + \mathbf{M}_{\Delta}^\mathcal{B}
\end{equation}
A first-order perturbation is used to approximate the perturbation:
\begin{equation}
    \mathbf{F}_{\Delta}^\mathcal{B} = \frac{\partial f(\mathbf{u})}{\partial \mathbf{u}}\vert_{\mathbf{x}_0, \mathbf{u}_0}\Delta\mathbf{u}, \quad 
    \mathbf{M}_{\Delta}^\mathcal{B} = \frac{\partial g(\mathbf{u})}{\partial \mathbf{u}}\vert_{\mathbf{x}_0, \mathbf{u}_0}\Delta\mathbf{u}
\end{equation}
where $\mathbf{x}_0$, $\mathbf{u}_0$ represent the trimmed state and the input.
Therefore, 
\begin{equation}\label{eq:wrenchconstraints}
    \mathbf{W}^\mathcal{B} = \left[ \begin{array}{c}
                        \mathbf{M}_{\Delta}^\mathcal{B} \\
                        \cdots\cdots \\
                        \mathbf{F}_{\Delta}^\mathcal{B}
                \end{array}\right] = \mathbf{A}\Delta\mathbf{u}
\end{equation}
with 
\begin{equation}
    \mathbf{A} = \left[ \begin{array}{c}
                        \frac{\partial g(\mathbf{u})}{\partial \mathbf{u}}\vert_{\mathbf{x}_0, \mathbf{u}_0} \\
                        \cdots\cdots\cdots \\
                        \frac{\partial f(\mathbf{u})}{\partial \mathbf{u}}\vert_{\mathbf{x}_0, \mathbf{u}_0}
                    \end{array}\right]
\end{equation}
as the control allocation's \textit{effectiveness matrix}. For the tiltrotor VTOL UAV with more actuators than required, the system is overdetermined with infinite number of solutions. 
The least-norm solution minimizes the total actuator effort while satisfying the force and moment constraints (see Equation~\ref{eq:wrenchconstraints}). It can be computed by the pseudo-inverse of $\mathbf{A}$. Actuator outputs that exist in the null space of $\mathbf{A}$ are utilized to satisfy other constraints, such as the actuator limits to avoid saturation. Therefore, 
\begin{equation}
    \Delta\mathbf{u} = \Delta\mathbf{u}_{\mathrm{LN}} +  \Delta\mathbf{u}_{\mathrm{Null}}
\end{equation}
where 
\begin{equation}
    \Delta\mathbf{u}_{\mathrm{Null}} = \Tilde{\mathbf{A}}\boldsymbol{\lambda}
\end{equation}
Here, $\mathbf{A}\Delta\mathbf{u}_{\mathrm{Null}} = 0$. $\Tilde{\mathbf{A}}$ is the matrix which columns are the basis of the null space of $\mathbf{A}$. $\boldsymbol{\lambda}$ is the coefficient of $\Tilde{\mathbf{A}}$, which will be solved by the following optimization problem: 
\begin{align} \label{eq:optimization}
    \min_{\boldsymbol{\lambda}} & J(\mathbf{u}_{\mathrm{sp}}) \\
    \mathrm{s.t} ~~&\mathbf{u}_{\mathrm{min},i}  \leq \mathbf{u}_{\mathrm{sp},i} \leq \mathbf{u}_{\mathrm{max},i} \nonumber
\end{align}
where $\mathbf{u}_{\mathrm{sp}}$ is the overall control setpoint:
\begin{equation}
    \mathbf{u}_{\mathrm{sp}} = \mathbf{u}_{0} + \Delta\mathbf{u}
\end{equation}
$J(\cdot)$ is the objective function where we try to minimize actuator change from the trimmed condition, 
\begin{equation}
    J = (\mathbf{u}_{\mathrm{sp}} - \mathbf{u}_{\mathrm{sp}, \mathrm{trim}})^\top\mathbf{R}(\mathbf{u}_{\mathrm{sp}}-\mathbf{u}_{\mathrm{sp}, \mathrm{trim}})
\end{equation}
where $\mathbf{R}$ is the weight matrix that accounts for the contribution from different actuators. The inequality constraints of the problem (see Equation~\ref{eq:optimization}) ensure that the output is within the actuators' limits, where $i$ represents the $\nth{i}$ actuator.

\section{Control in Actuator Failure} \label{sec:fail-control}
In this section, we analyze the effect of a set of actuator failure cases (see Table.~\ref{tab:failcase}). We then propose changes to the nominal control to enable the system's recovery from failures. 

\begin{table}
    \centering
    \caption{Actuator failure cases in different VTOL flight phases.}
    \label{tab:failcase} 
    \begin{adjustbox}{width=0.48\textwidth}
    \begin{threeparttable}
    \begin{tabular}{cll}
    \hline\hline
    Flight Phase                & Failure Description            & Cause                      \\
    \hline\hline
    \multirow{2}{*}{Multirotor} & Lock of one tilt in hover      & Broken servo               \\
                                & Single motor failure in hover  & Motor flaw/propeller loss  \\
    \hline
    \multirow{4}{*}{Fixed-wing}   & Single motor failure in cruise & Motor flaw/propeller loss   \\
                                & Lock of one elevator in cruise & Broken servo               \\
                                & Lock of one aileron in cruise  & Broken servo  \\
    \hline\hline
    \end{tabular}
        \begin{tablenotes}[normal,flushleft]
          \item $^\dagger$ We consider the failures in the multirotor and fixed-wing cruise flight phases since most of the flight stays in those two phases.
        \end{tablenotes}
    \end{threeparttable}
    \end{adjustbox}
    \vspace{-5mm}
\end{table}


For the tiltrotor VTOL designed in Sec.\ref{sec:system}, there are 12 actuators to control the six degrees of freedom (6~DOF) rigid body motion. The system, therefore, has 6~DOF control redundancy. This redundancy enables the system to handle the actuator failure. For example, suppose one of the motors fails in the multirotor phase. In that case, the tilts can be active and compensate for this failure with different rotor speeds and thrust directions, enabling the aircraft to maintain its flight. Alternatively, suppose the elevator gets locked during the fixed-wing cruise flight. In that case, although the aircraft loses some degree of pitch control authority for the fixed-wing configuration, the combination of multirotor and tilts can compensate for it. Therefore, we adjust our nominal control allocation to adapt to the actuator failure as long as the control wrench is within the feasible wrench space discussed in Sec.~\ref{sec:wrench-space}.

We denote by $^k\mathbf{A}$ the control effectiveness matrix in which the $\nth{k}$ column of $\mathbf{A}$ has been zeroed or equivalently, removed. Such matrix represents the control effectiveness matrix of the tiltrotor VTOL in which the $\nth{k}$ actuator does not work anymore after a failure. In other words, $\mathbf{u}_k = u_f$ for the tilt lock or control surface (aileron or elevator) lock scenarios, where $u_f$ is a constant value within the servo limit. For motor failure, $u_f$ is usually a small value or zero. Here, we assume that $u_f$ is known.\footnote{~Some failure detection and identification mechanisms can be deployed on various aircraft types to achieve this~\cite{Keipour:2019:icra:anomaly, Keipour:2021:ijrr:alfa}.} $\mathbf{u}_{i\neq k}$ is the actuator input for all the functioning actuators. We also denote $\,^k\mathbf{a}$ the $\nth{k}$ column of $\mathbf{A}$. Thus, we have the desired control wrench
\begin{align}
    \mathbf{W}^{\mathcal{B}} &= \mathbf{W}_{i\neq k}^{\mathcal{B}} + \mathbf{W}_k^{\mathcal{B}} = \,^k\mathbf{A}\mathbf{u}_{i\neq k} + \,^k\mathbf{a}\mathbf{u}_k \\
    \mathbf{W}_{i\neq k}^{\mathcal{B}} &= \,^k\mathbf{A}\mathbf{u}_{\mathrm{trim},i\neq k} +  \,^k\mathbf{A} \Delta \mathbf{u}_{i\neq k} \nonumber\\
                         &=\mathbf{W}_{\mathrm{trim}, i\neq k}^{\mathcal{B}} + \mathbf{W}_{\Delta, i\neq k}^{\mathcal{B}}\nonumber
\end{align}
where $\mathbf{W}_{i\neq k}^{\mathcal{B}}$ and $\mathbf{W}_k^{\mathcal{B}}$ represent the control wrench generated by the normal and failed actuators. 
Therefore,
\begin{equation}\label{eq:failpro}
    \mathbf{W}^{\mathcal{B}} - \mathbf{W}_{k}^{\mathcal{B}} - \mathbf{W}_{\mathrm{trim}, i\neq k}^{\mathcal{B}} = \,^k\mathbf{A}\Delta \mathbf{u}_{i\neq k}
\end{equation}

Note that Equation~\eqref{eq:failpro} has the same format as Equation~\eqref{eq:wrenchconstraints}. Therefore, we solve the constraint optimization problem to find the output for the normal actuators. 

\section{Experiments and Results} \label{sec:tests}

\subsection{Experiment Setup}
We modeled the tiltrotor VTOL in the Gazebo simulator based on the design described in Sec.~\ref{sec:system}. The dynamic control allocation is developed on top of the PX4 source code, which can run directly on real aircraft. The constrained optimization is solved using Algilib~\cite{Algilib}, which is an open-source header-only numerical analysis and data processing library. The numerical values of the constants used during the experiment are shown in Table.~\ref{tab:parameters}.

\begin{table}
    \centering
    \caption{Physical parameters of the aircraft.}
    \label{tab:parameters} 
    \begin{adjustbox}{width=0.48\textwidth}
    \begin{tabular}{lll}
    \hline\hline
    Parameter                & Description            & Value           \\
    \hline\hline
    $\mathrm{m}$             & Mass                         & 4.6 kg               \\
    $b$                      & Wingspan                       & 2.0 m               \\
    $\rho$                   & Air density                    & 1.2250 $\mathrm{kg/m^3}$  \\
    $\bar{c}$                & Mean chord                     & 0.22 m  \\
    $S$                      & Wing surface area              & 0.44 $\mathrm{m}^2$               \\
    $c_T$                  & Propeller thrust coefficient   & 2.2164e-5               \\
    $c_K$                  & Propeller torque coefficient   & 1.1082e-6  \\
    $C_{La}$                 & Aileron coefficient            & 0.1173  \\
    $C_{Me}$                 & Elevator coefficient           & 0.5560  \\
    $C_{Nr}$                 & Rudder coefficient             & 0.0881  \\
    $C_{Z, 0}$, $C_{Z, \alpha}$        & Lift coefficient             & 0.35, 0.11  \\
    $C_{D, 0}$, $C_{D, \alpha}$        & Drag coefficient             & 0.01, 0.2  \\
    \hline\hline
    \end{tabular}
    \end{adjustbox}
\end{table}



\subsection{Results}
\subsubsection{Motor Failure in Hover}
\begin{figure}
    \centering
    \includegraphics[width=\linewidth]{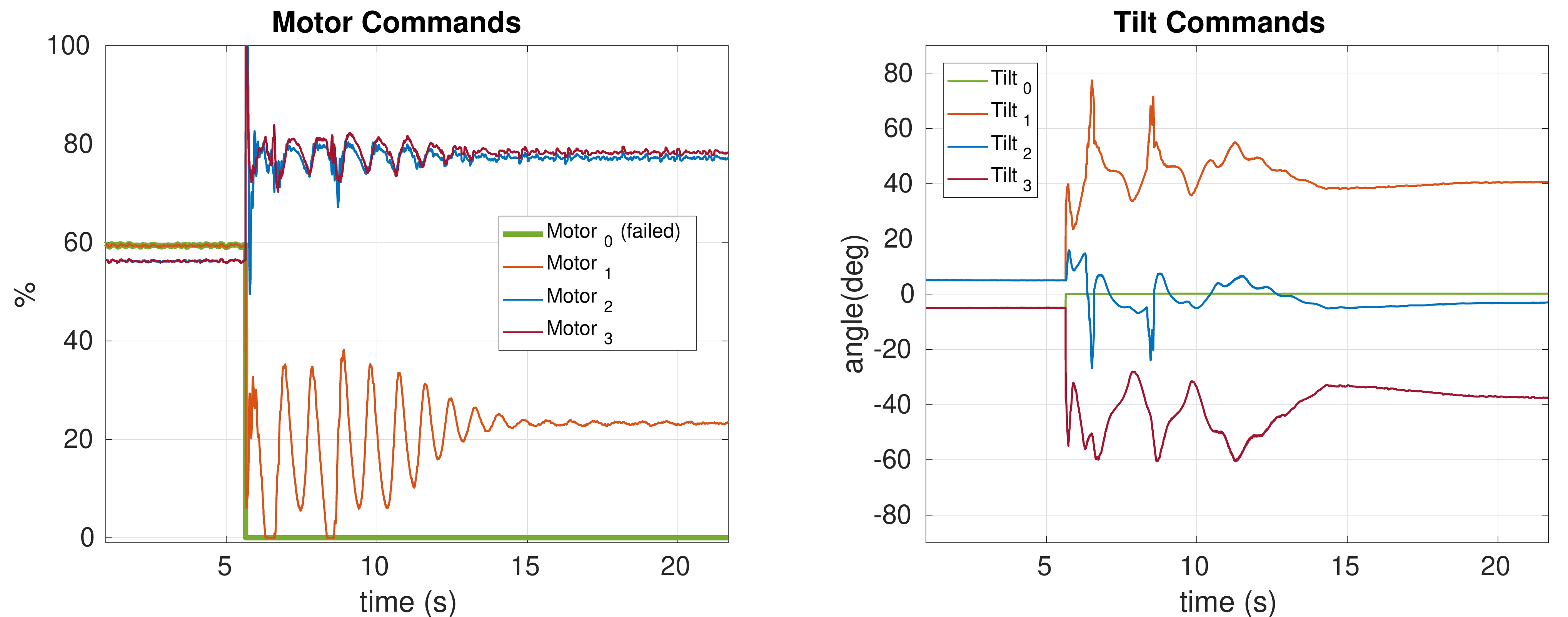}
    \caption{Actuator commands when a motor failure happens in hover.}
    \label{fig:motor_failure}
    \vspace{-5mm}
\end{figure}

To test this failure case, we completely shut down one of the motors after the take-off during hovering and before the transition. Figure~\ref{fig:motor_failure} shows the allocated actuator commands. It is clear that when the front right motor fails, all the functioning actuators adjust themselves and compensate for the thrust loss. Notice that the tilt angle corresponding to the failed motor quickly converges to zero, which is reasonable since the failed rotor can no longer generate any effective wrench. The actuator commands converge after about 10 seconds. After recovery, the aircraft is still controllable and can follow the desired waypoints.  

\begin{figure}
    \centering
    \includegraphics[width=\linewidth]{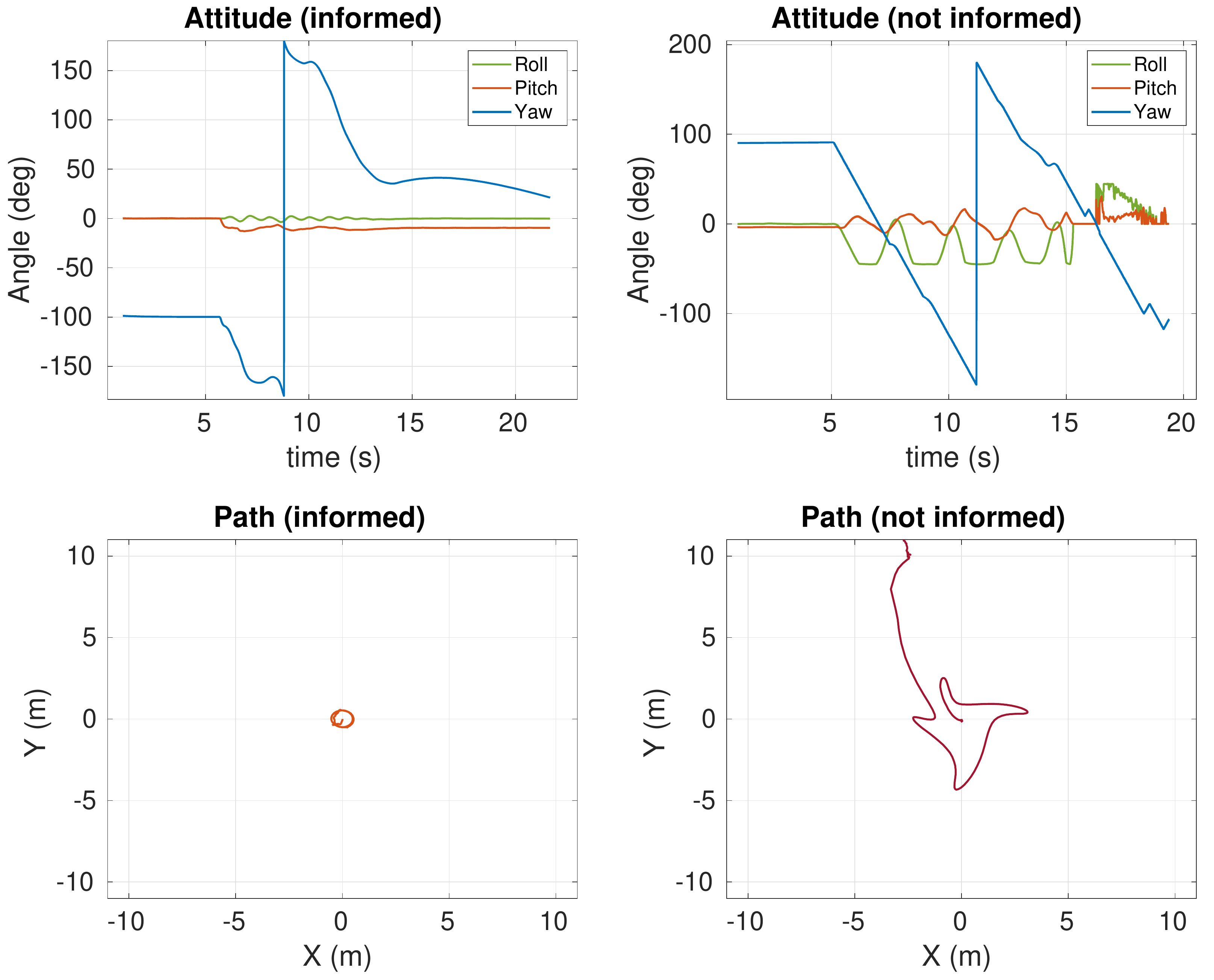}
    \caption{Aircraft attitude and path when a motor failure happens in hover.}
    \label{fig:attitude_path}
\end{figure}

We compare our method with the case where the controller is not informed about the motor failure. Figure~\ref{fig:attitude_path} presents the aircraft attitude and flight path for the two scenarios: with and without being informed of the system failure. Without the failure knowledge, the system is still trying to allocate a control wrench to the failed actuator, resulting in the aircraft immediately getting into an aggressive rotation, losing control, and crashing. 

\subsubsection{Tilt Angle Failure in Hover}
\begin{figure}
    \centering
    \includegraphics[width=\linewidth]{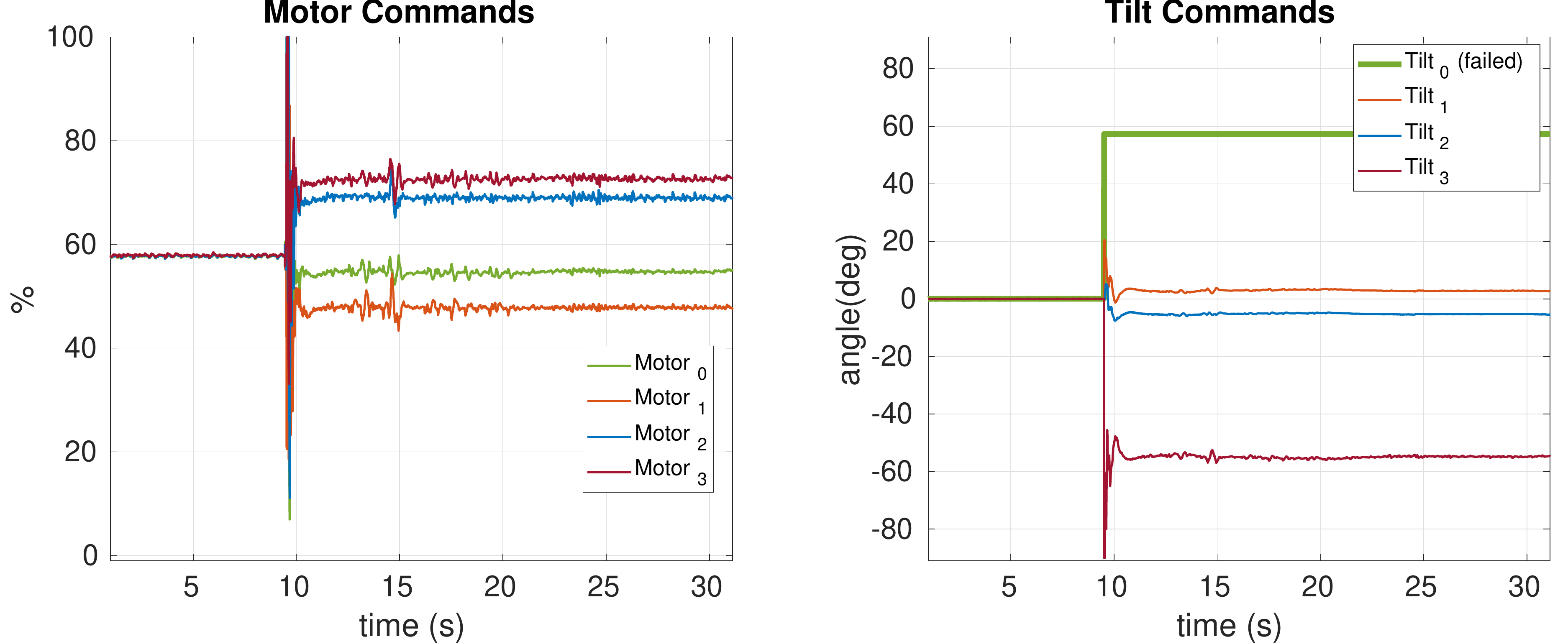}
    \caption{Actuator commands when a tilt servo failure happens in hover.}
    \label{fig:tilt_failure}
    \vspace{-4mm}
\end{figure}

This failure was simulated by suddenly locking the tilt servo at a fixed position (here at about 60\degree~tilt). Figure~\ref{fig:tilt_failure} shows the actuator commands. We can see that even with a 60\degree~tilt angle change, the system can quickly recover after a significant sudden disturbance by dynamically reallocating to the rest of the actuators. Successful flights were performed with $\mathcal{X}_i = \pm 90~\mathrm{deg}$.

\subsubsection{Motor Failure During Cruise Flight}
\begin{figure}
    \centering
    \includegraphics[width=\linewidth]{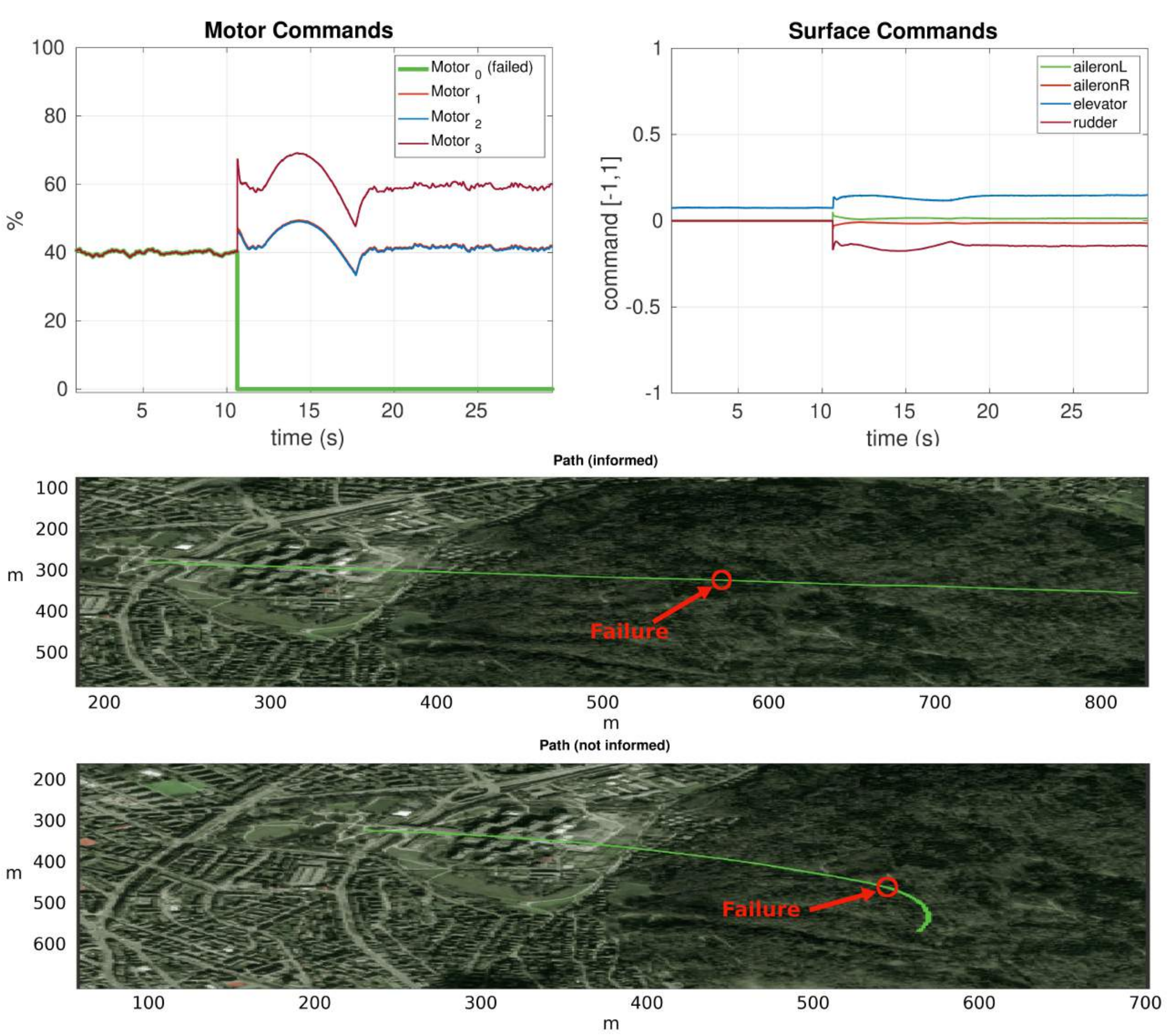}
    \caption{Actuator commands and the flight path when a motor failure happens in cruise flight.}
    \label{fig:fw_motor_failure}
\end{figure}

Figure~\ref{fig:fw_motor_failure} shows a simulated motor failure during cruise flight. The dynamical control allocation enables the aircraft to adapt to the failure and quickly return to the normal flight. It is worth noting that the optimization solution results in all the tilts being kept at full tilts (all the rotors are facing forward). By adjusting the motor speeds and the control surfaces, the vehicle can recover from failure. In fact, in high-speed cruise flight, relying more on motor speeds and control surfaces positively affects the aircraft since the drastic tilt change may be hazardous to the aircraft structure. Compared to the scenario of the system without the failure knowledge, it is clear that being aware of the failure allows the aircraft to maintain a straight path while being agnostic to the failure leads to a significant path deviation. 

\subsubsection{Elevator Failure During Cruise Flight}
\begin{figure}
    \centering
    \includegraphics[width=\linewidth]{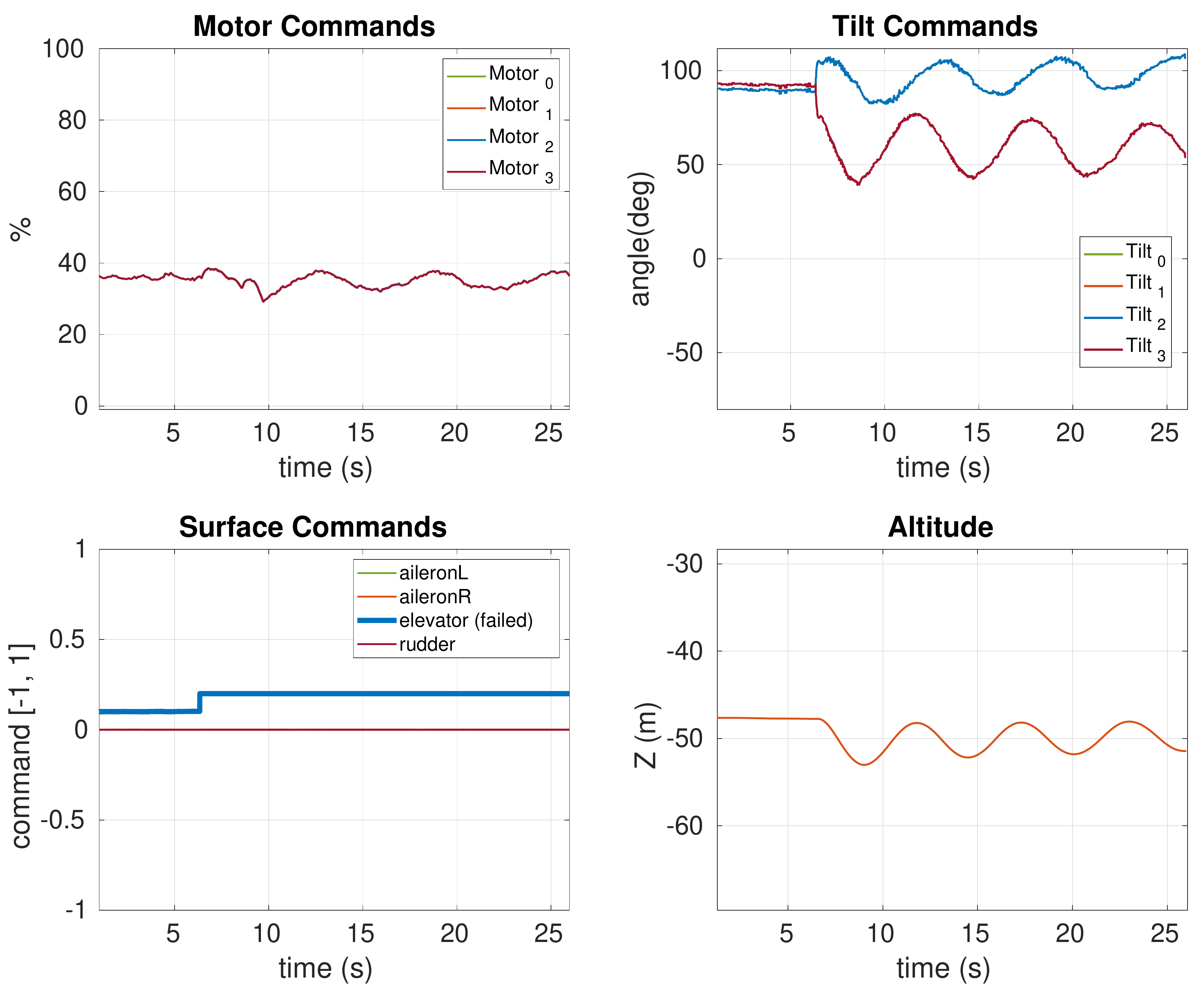}
    \caption{Actuator commands and the flight altitude when the elevator failure happens in cruise flight.}
    \label{fig:fw_elevator_failure}
    \vspace{-4mm}
\end{figure}

We inject the elevator failure with the elevator locked at 6\degree. Since the elevator is the only control surface for pitch, the tilt angles must adapt when failure happens. Fig.~\ref{fig:fw_elevator_failure} shows the allocated actuator commands. The rear two rotors tilt back, and the front two tilt slightly forward over 90\degree~to compensate for the constant pitch-up moment generated by the locked elevator. The aircraft can keep a reasonable constant altitude with only a 4~m variation. 

\subsubsection{Aileron Failure During Cruise Flight}
\begin{figure}
    \centering
    \includegraphics[width=0.8\linewidth]{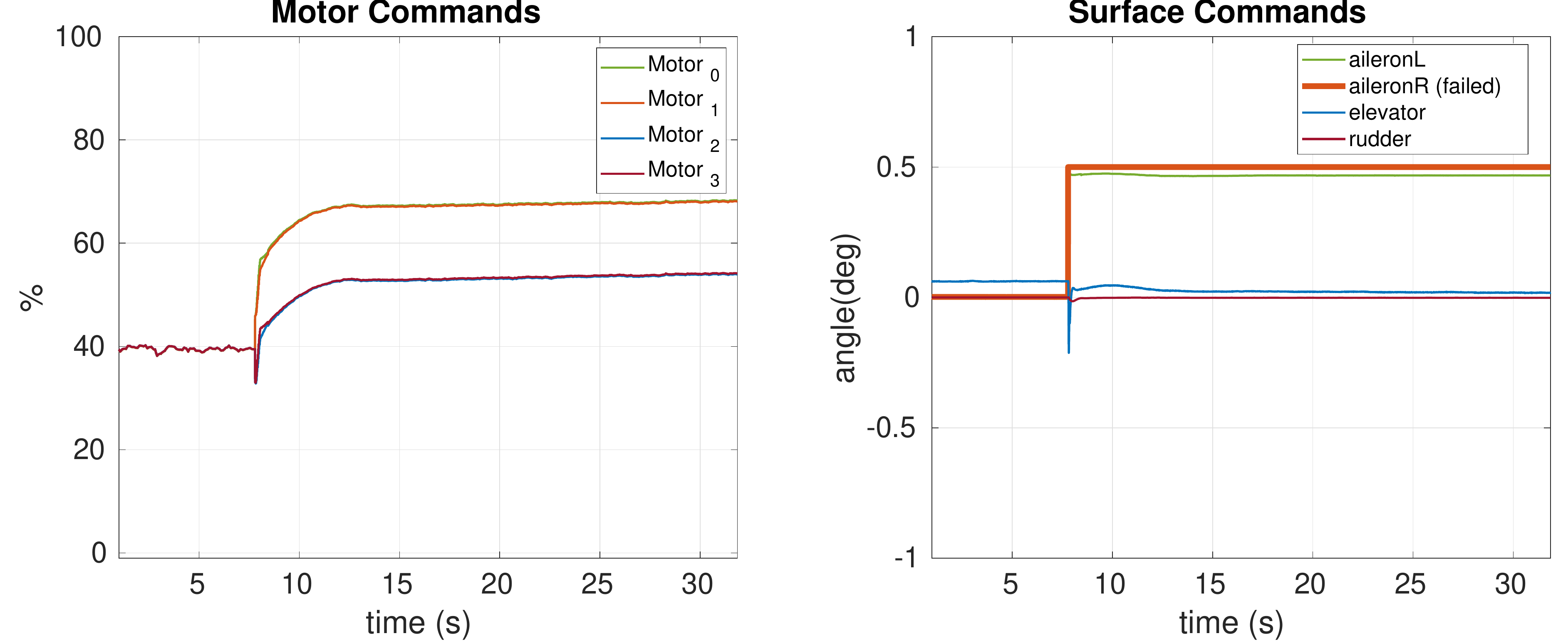}
    \caption{Actuator commands when aileron fails in cruise flight.}
    \label{fig:fw_aileron_failure}
    \vspace{-6mm}
\end{figure}

When the aileron gets locked at 15\degree, the other healthy control surfaces adjust to compensate for the lost roll control authority (see Figure~\ref{fig:fw_aileron_failure}). In particular, the other aileron has a significant adaptation to compensate for the roll. At the same time, the motors speed up with the left two spinning faster than the right two to compensate for the induced yaw moment due to the sudden sideslip caused by the roll disturbance. The tilt angles still keep facing forward to avoid drastic structure change. The controller can handle the locked aileron in calm wind conditions up to $\delta_a = \pm 15\degree$, which is about half of the maximum aileron deflection range.

\section{Conclusion and Future Work} \label{sec:conclusion}

This paper introduced fault tolerance to the design of a tiltrotor VTOL. The VTOL is a hybrid of a fixed-wing aircraft and a variable-pitch quadrotor. The rotor arms' design with airfoil cross-sections decouples the structure of the quadrotor arm and the main wing, making the tilt design cleaner and the control less affected by the unexpected wing structure deformation. Each propeller can rotate individually. 

We modeled the aircraft's nonlinear dynamics and analyzed the feasible wrench space that the vehicle can generate. Then, we designed the dynamic control allocation so that the system can adapt to actuator failures. The proposed approach is lightweight and is implemented as an extension to the PX4 flight control stack. 

Finally, we presented extensive experiment results validating the system's performance under different actuator failures. Future research includes performing the real flight test to validate the developed approach further.







\bibliographystyle{IEEEtran}
\bibliography{paper-citations.bib}

\end{document}